# A Knowledge Discovery Framework for Learning Task Models from User Interactions in Intelligent Tutoring Systems

Philippe Fournier-Viger[1], Roger Nkambou[1], and Engelbert Mephu Nguifo[2]

[1] University of Quebec in Montreal, Montreal (QC), Canada
[2] Université Lille-Nord de France, Artois, F-62307 Lens, CRIL, F-62307 Lens
CNRS UMR 8188, F-62307 Lens, France
fournier_viger.philippe@courrier.uqam.ca, nkambou.roger@uqam.ca, mephu@cril.univ-artois.fr

**Abstract.** Domain experts should provide relevant domain knowledge to an Intelligent Tutoring System (ITS) so that it can guide a learner during problem-solving learning activities. However, for many ill-defined domains, the domain knowledge is hard to define explicitly. In previous works, we showed how sequential pattern mining can be used to extract a partial problem space from logged user interactions, and how it can support tutoring services during problem-solving exercises. This article describes an extension of this approach to extract a problem space that is richer and more adapted for supporting tutoring services. We combined sequential pattern mining with (1) dimensional pattern mining (2) time intervals, (3) the automatic clustering of valued actions and (4) closed sequences mining. Some tutoring services have been implemented and an experiment has been conducted in a tutoring system

## 1 Introduction

Domain experts should provide relevant domain knowledge to an Intelligent Tutoring System (ITS) so that it can guide a learner during problem-solving activities. One common way of acquiring such knowledge is to use the method of cognitive task analysis that aims at producing effective problem spaces or task models by observing expert and novice users for capturing different ways of solving problems. However, cognitive task analysis is a very time-consuming process [1] and it is not always possible to define a satisfying complete or partial task model, in particular when a problem is ill-structured. According to Simon [2], an ill-structured problem is one that is complex, with indefinite starting points, multiple and arguable solutions, or unclear strategies for finding solutions. Domains that include such problems and in which, tutoring targets the development of problem-solving skills are said to be ill-defined (within the meaning of Ashley et al. [3]). An alternative to cognitive task analysis is constraint-based modeling (CBM) [4], which consist of specifying sets of constraints on what is a correct behavior, instead of providing a complete task description. Though this approach was shown to be effective for some ill-defined domains, a domain expert has to design and select the constraints carefully.

Contrarily to these approaches where domain experts have to provide the domain knowledge, a promising approach is to use knowledge discovery techniques for automatically learning a partial problem space from logged user interactions in an ITS, and to use this knowledge base to offer tutoring services.

We did a first work in this direction [5] by proposing a framework to learn a knowledge base from user interactions in procedural and ill-defined domains [5]. The framework takes as input sequences of user actions performed by expert, intermediate and novice users, and consist of applying two knowledge discovery techniques. First, sequential pattern mining (SPM) is applied to discover frequent action sequences. Then, association rules discovery find associations between these significant action sequences, relating them together. The framework was applied in a tutoring system to extract a partial problem space that is used to guide users, and thus showed to be a viable alternative to the specification of a problem-space by hand for the same domain [5, 7]. This framework differs from other works that attempt to construct a task model from logged student interactions such as [8], [9] and [10], since these latter are devoid of learning, reducing these approaches to simple ways of storing or integrating raw user solutions into structures.

Although the framework [5] was shown to be useful, it can be improved in different ways. Particularly, in this paper, we present an extended SPM algorithm for extracting a problem space that is richer and more adapted for supporting tutoring services. This work was done following our application of the framework in the RomanTutor tutoring system [5].

The rest of the paper is organized as follow. First, it introduces RomanTutor [6] and the problem of SPM from user actions. Then it presents the limitations of the framework encountered and extensions to address these issues. Finally, it presents preliminary results of the application of the improved framework in RomanTutor, future work and a conclusion.

## 2 The RomanTutor Tutoring System

RomanTutor [6] (cf. fig. 1) is a simulation-based tutoring system to teach astronauts how to operate Canadarm2, a 7 degrees of freedom robot manipulator deployed on the International Space Station (ISS). During the robot manipulation, operators do not have a direct view of the scene of operation on the ISS and must rely on cameras mounted on the manipulator and at strategic places in the environment where it operates. The main learning activity in RomanTutor is to move the arm to a goal configuration. To perform this task, an operator must select at every moment the best cameras for viewing the scene of operation among several cameras mounted on the manipulator and on the space station.

In previous work [7], we attempted to model the Canadarm2 manipulation task with a rule-based knowledge representation model. Although, we described high-level rules such as to set the parameters of cameras in a given order, it was not possible to go in finer details to model how to rotate the arm joint(s) to attain a goal configuration. The reason is that for a given robot manipulation problem, there are many possibilities for moving the robot to a goal configuration and thus, it is not possible to define a complete and explicit task model. In fact there is no simple ‘legal move generator’ for finding all the possibilities at each step. Hence, RomanTutor operates in an ill-defined-domain. As a solution, we identified 155 actions that a learner can take, which are (1) selecting a camera, (2) performing a small/medium/big increase or decrease of the pan/tilt/zoom of a camera and (3) applying a small/medium/big positive/negative rotation value to an arm joint. Then, we applied SPM to mine frequent action sequences from logged users' interactions [5]. The resulting knowledge base served in RomanTutor to track the patterns that a learner follows, and to suggest the next most probable actions that one should execute.

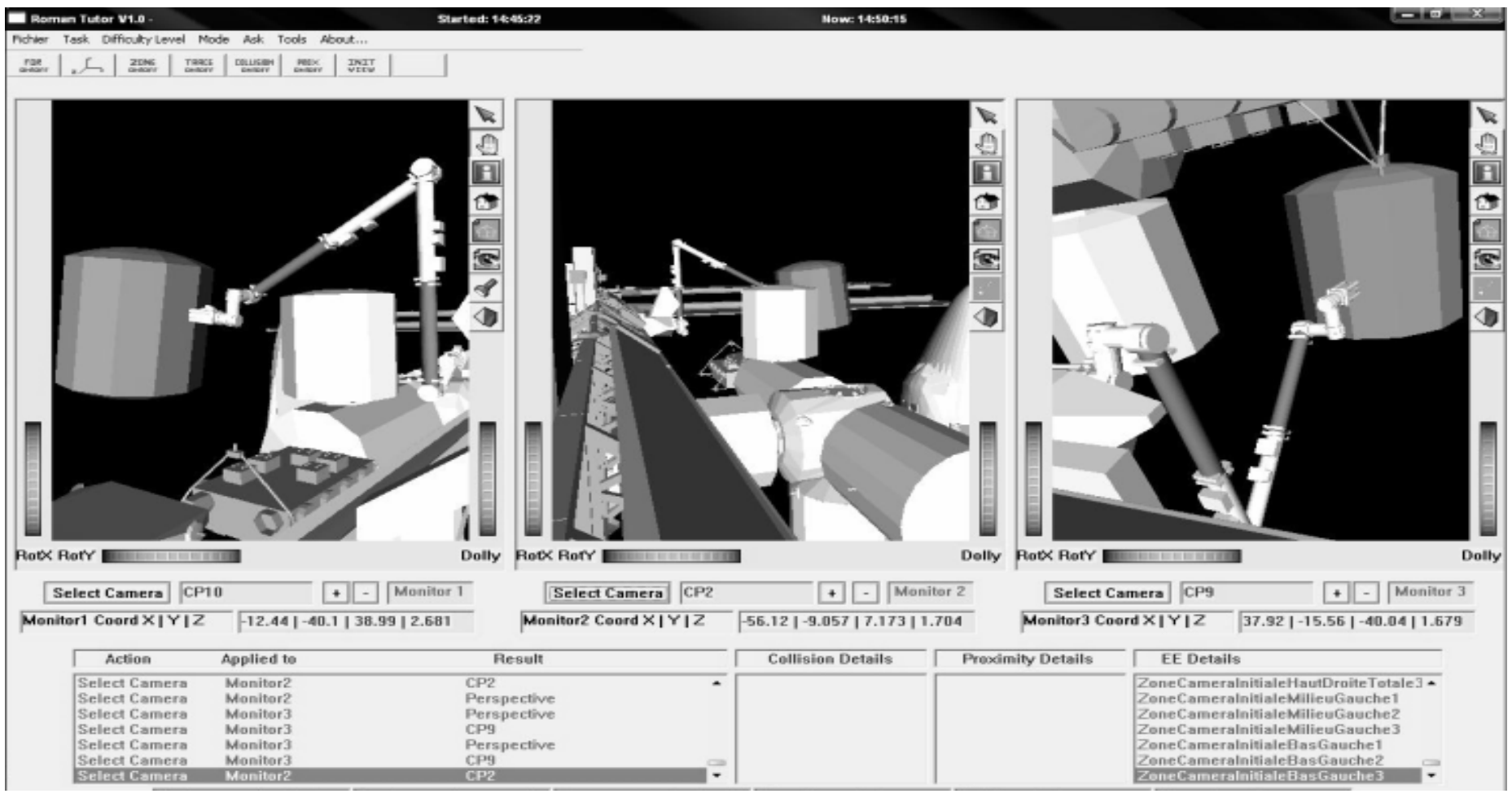

**Fig. 1.** The RomanTutor User Interface

## 3 Sequential Patterns Mining from User Actions

The problem of mining sequential patterns is stated as follows [11]. Let D be a transactional database containing a set of transactions (here also called plans) and a set of sequence of items (here called actions). An example of D is depicted in figure 2.a. Let $A = \{a_1, a_2, \ldots, a_n\}$ be a set of actions. We call a subset $X \subseteq A$ an actionset and $|X|$, its size. Each action in an actionset (enclosed by curly brackets) are considered simultaneous. A sequence $s = (X_1, X_2, \ldots, X_m)$ is an ordered list of actionsets, where $X_i \subseteq A$, $i \in \{1,\ldots,m\}$, and where m is the size of s (also noted $|s|$). A sequence $s_a = (A_1, A_2, \ldots, A_n)$ is contained in another sequence $s_b = (B_1, B_2, \ldots, B_m)$ if there exists integers $1 \leq i_1 < i_2 < \ldots < i_n \leq m$ such that $A_1 \subseteq B_{i1}$ , $A_2 \subseteq B_{i2}$ , . . . , $A_n \subseteq B_{in}$. The relative support of a sequence $s_a$ is defined as the percentage of sequences $s \in D$ that contains $s_a$, and is denoted by $supD(s_a)$. The problem of mining sequential patterns is to find all the sequences $s_a$ such that $supD(s_a) \geq minsup$ for a database D, given a support threshold *minsup*.

Consider the dataset of figure 2.a. The size of the plan 2 is 6. Suppose we want to find the support of S2. From figure 2.a, we know that S2 is contained in plan 1, 2 and 5. Hence, its support is 3 (out of a possible 6), or 0.50. If the user-defined minimum support value is less than 0.50, then S2 is deemed frequent. To mine sequential patterns several algorithms have been proposed [11, 12, 13]. In our first experiment in RomanTutor, we chose PrefixSpan [12] as it is a promising approach for mining large sequence databases having numerous patterns and/or long patterns, and also because it can be extended to mine sequential patterns with user-specified constraints. Figure 2.b shows some sequential patterns extracted by PrefixSpan from the data in figure 2.a using a minimum support of 25%. In RomanTutor, one mined pattern is for example, to select the camera 6, which gives a close view of the arm in its initial position, slightly decrease the yaw of camera 6, select the elbow joint and decrease a little bit its rotation value. Although, the set of patterns extracted for RomanTutor constitutes a useful problem space that capture different ways of solving problems, we present next the limitations of SPM encountered in our first experiment, and extensions to PrefixSpan to address these issues.

| ID | Sequences of actions |
|---|---|
| 1 | 1 2 25 46 48 {9 10 11 31} |
| 2 | 1 25 46 54 {10 11 25} 48 |
| 3 | 1 2 3 {9 10 11 31} 48 |
| 4 | 2 3 25 46 11 {14 15 48} 74 |
| 5 | 4 1 25 27 46 48 |
| 6 | 1 3 44 45 46 48 |

→

| ID | Seq. patterns | Support |
|---|---|---|
| S1 | 1 46 48 | 66 % |
| S2 | 1 25 46 48 | 50 % |
| S3 | 1 25 46 {10 11} | 33 % |
| S4 | 1 {9 10 31} | 33 % |
| S5 | 1 {9 11 31} | 33 % |
| … | … | … |

**Fig. 2.** (a) A Data Set of 6 Plans (b) Example of Sequential Patterns Extracted

## 4 Extending Sequential Pattern Mining with Time Intervals

A first limitation that we encountered is that extracted patterns often contain "gaps" with respect to their containing sequences. For instance, in the example of figure 2, action "2" of plan 1 has not been kept in S1. A gap of a few actions is ok in a tutoring context because it eliminates non-frequent learners' actions. But when a sequence contain many or large gap(s), it becomes difficult to use this sequence to track a learner's actions and to suggest a next relevant step. Thus, there should be a way of limiting the size of the gaps in mined sequences. Another concern is that some patterns are too short to be useful in a tutoring context (for example, sequences of size 1). In fact, there should be a way of specifying a minimum sequential pattern size.

An extension of SPM that overcomes these limitations is to mine patterns from a database with time information. A time-extended database is defined as a set of time-extended sequences s = $<(t_1,X_1), (t_2,X_2),\ldots, (t_n,X_n)>$, where each actionset $X_x$ is annotated with a timestamp tx. Each timestamp represents the time elapsed since the first actionset of the sequence. Actions within a same actionset are considered simultaneous. For example, one time-extended sequence could be <(0, a), (1, b c), (2, d)>, where action d was done one time unit after b and c, and two time units after a. The time interval between two actionsets $(t_x,X_x)$ and $(t_y,X_y)$ is calculated as $|t_x - t_y|$. In this work, we suppose a time interval of one time unit between any adjacent actionsets, so that $|t_x - t_y|$ become a measure of the number of actionsets between $(t_x, X_x)$ and $(t_y, X_y)$. The problem of Generalized Sequential Pattern Mining with Time Intervals (GSPM) [14] is to extract all time-extended sequences s from a time-extended database, such that supD(s) ≥ minsup and that s respect all time constraints. Four types of constraints are proposed by Hirate and Yamana [14]. The constraints C1 and C2 are the minimum and maximum time interval required between two adjacent actionsets of a sequence (gap size). The constraints C3 and C4 are the minimum and maximum time interval required between the head and tail of a sequence. For example, for the sequence <(0, a), (1, b c), (2, d)>, the time interval between the head and the tail is 2 and the time interval between the actionset (0, a) and (1, b c) is 1.

Hirate and Yamana [14] have proposed an extension of the PrefixSpan algorithm for the problem of GPSM. We present it below –with slight modifications- as it is the basis of our work. The algorithm finds all frequent time-extended sequences in a database ISDB that respect minsup, C1, C2, C3 and C4, by performing a depth-first search. The algorithm is based on the property that if a sequence is not frequent, any sequence containing that sequence will not be frequent. The algorithm proceeds by recursively projecting a database into a set of smaller projected databases. This process allows growing patterns one action at a time by finding locally frequents actions.

In the following, the notation ISDB|(t,i) represents the time-extended database resulting from the operation of projecting a time-extended database ISDB with a pair (timestamp, item). ISDB|(t, i) is calculated as follow.

```
ISDB((t,i))
  ISDB|(t,i) := Ø.
  FOR each sequence σ=<(t1,X1),(t2,X2)…(tn,Xn)> of
ISDB.
  FOR each actionset (tx,Xx) of σ containing i.
    IF Xx/{i} = Ø
      s :=<(tx+1-tx,ax+1), … (tn-tx,Xn)>
    ELSE
      s :=<(0, Xx/{i}), (tx+1-tx, ax+1), … (tn-tx, Xn)>
    IF s ≠ Ø  and s satisfies C1, C2 and C4
      Add s to ISDB|(t,i).
  Return ISDB|(t,i).
```

The Hirate-Yamana algorithm (described below) discovers all frequent time-extended sequences.

```
algoHirate(ISDB, minsup, C1, C2, C3, C4)

  R := ∅.

  Scan ISDB and find all frequent items with support
   higher than minsup.
  FOR each frequent item i,
    Add (0, i) to R.
    algoProjection(ISDB|(0, i), R, minsup,C1,C2,C3,C4).

  RETURN R;

  algoProjection(ISDB|prefix, R, minsup, C1, C2, C3, C4)
    Scan ISDB|prefix to find all pairs of item and
     timestamp, denoted (t, i) satisfying minsup, C1 and
  C2.
```

```
    FOR each pair (t, i) found
      newPrefix := Concatenate(prefix, (t, i)).
      IF newPrefix satisfies C3 and C4
        Add newPrefix to R.
        IF (size of ISDB|newPrefix) >= minsup
         algoProjection(ISDB|newPrefix, R, minsup,C1,C2
  C3,C4).
```

To illustrate the Hirate-Yamana algorithm, let's consider applying it to the database ISDB depicted in figure 3, with a *minsup* of 50 % and the constraint C2 equals to 2 time units. In the first part of *algoHirate*, frequent actions *a*, *b* and *c* are found. As a result <(0,a)>, <(0,b)> and <(0,c)> are added to *R*, the set of sequences found. Then, for *a*, *b* and *c*, the projected database ISDB|(0,a), ISDB|(0,b) and ISDB|(0,c) are created, respectively. For each of these databases, the algorithm *algoProjection* is executed. *algoProjection* first finds all frequent pairs (timestamp, item) that verify the constraints C1, C2 and *minsup*. For example, for ISDB| (0,a), the frequent pair (0,b) and (2,a) are found. These pairs are concatenated to (0,a) to obtain sequences <(0,a), (0,b)> and <(0,a), (2,a)>, respectively. Because these sequences respect C3 and C4, they are added to the set R of sequences found. Then, the projected database ISDB|<(0,a), (0,b)> and ISDB|<(0,a), (2,a)> are calculated. Because these databases contain more than *minsup* sequences, *algoProjection* is executed again. After completing the execution of the algorithm, the set of sequences R contains <(0,a)>, <(0,a), (0,b)>, <(0,a), (2,a)>, <(0,b)> and <(0,c)>.

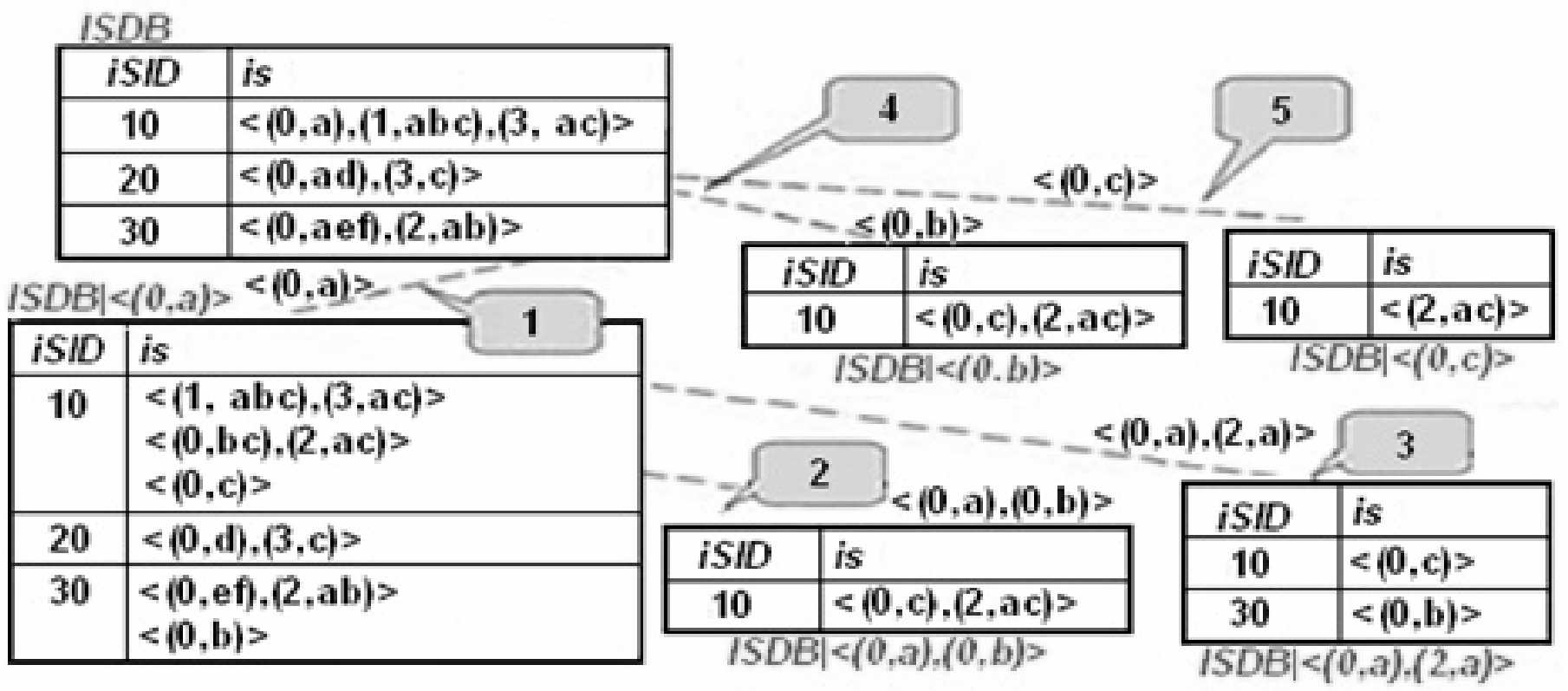

**Fig. 3.** An application of the Hirate-Yamana Algorithm (adapted from [10])

## 5 Extending SPM with Automatic Clustering of Valued Actions

A second limitation that we encountered when we applied PrefixSpan to extract a problem space is that it relies on a finite set of actions. As a consequence, if some actions were designed to have parameters or values, they have to be defined as one or more distinct actions. For example, in our first experiment in RomanTutor, we categorized the joint rotations as small, medium and big, which correspond respectively to 0 to 60, 60 to 100, and more than 140 degrees. The disadvantage with this categorization is that it is fixed. In order to have dynamic categories of actions, we extended the Hirate-Yamana algorithm to perform an automatic clustering of valued actions.

We propose to define a valued sequence database as a time-extended sequence database, where sequences can contain valued actions. A valued action *a{value}*, is defined as an action *a* that has a value *value*. In this work, we consider that a value is an integer. For example, the sequence <(0,a{2}), (1,b), (2,bc{4})> contains the valued action *a*, and *b* with values 2 and 4, respectively. The left part of figure 4 shows an example of a sequence database containing valued actions.

| ID | Time-extended sequences |
|---|---|
| 1 | <(0,a{2}), (1,bc{4})> |
| 2 | <(0,a{2}), (1,c{5}))> |
| 3 | <(0,a{5}), (1,c{6}))> |
| 4 | < (0,f), (1,a{6}))> |
| 5 | <(0, f b{3}), (1,e) , (2,f))> |
| 6 | <(0,b{2}), (1,d))> |

→

| Mined valued seq. patterns | Supp. |
|---|---|
| <(0,a{2})> | 33 % |
| <(0,a{5})> | 33 % |
| <(0,a{2}), (1, c{5})> | 33 % |
| <(0,c{5})> | 50 % |
| <(0,f)> | 33 % |
| … | … |

**Fig. 4.** Applying Hirate-Yamana with Automatic Clustering of Valued Actions

To mine patterns from a valued database, we added a special treatment for valued actions. We modified the action/pair counting of the Hirate-Yamana algorithm to note the values of the action being counted, and their sequence ids. We modified the database projection operation ISDB|(t,i) so that it can be called with a valued action *i* and a set of values $V=\{v_1, v_2 \ldots, v_n\}$. If the support of *i* in ISDB is higher or equals to 2 * *minsup*, the database projection operation calls the K-Means algorithm [15] to find clusters. The K-Means algorithm takes as parameter K, a number of clusters to be created, and the set of values V to be clustered. K-Means first creates K random clusters. Then it iteratively assigns each value from V to the cluster with the nearest median value until all clusters remain the same for two successive iterations. In the database projection operation, K-Means is executed several times starting with K=2, and incrementing K until the number of frequent clusters found (with size >= *minsup*) does not increase. This larger set of frequent clusters is kept. Then, the sequences of ISDB|(t,i) are separated into one or more databases according to these clusters. Afterward, *algoProjection* is called for each of these databases with size equal or greater than *minsup*.

Moreover, if ISDB|(t,i) is called from *algoHirate* and *n* clusters are found, instead of just adding <(o,{i})> to the set R of sequences found, <(0, i{$v_{x1}$})>, <(0, i{$v_{x2}$})> … <(0, i{$v_{xn}$})> are added, where $v_{x1}$, $v_{x2}$, .. $v_{xn}$ are the median value of each cluster. Similarly, we have adapted *algoProjection* so that sequences are grown by executing *Concatenate* with (t, i{$v_{x1}$}), (t, i{$v_{x2}$}) … (t, i{$v_{xn}$}, instead of only <(t,{i})>.

The right part of figure 4 shows some sequences obtained from the execution of the modified algorithm with a *minsup* of 32 % (2 sequences) on the valued sequence database depicted in the left part of figure 4. From this example, we can see that the action "a" is frequent (with a support of 4) and that two clusters were dynamically created from the set of values {2, 2, 5, 6} associated to "a". The first cluster contained the values 2 and 2 with a median value of 2 and the second one contained 5 and 6 with a median value of 5. This resulted in frequent sequences containing "a{2}"and some other sequences containing "a{5}".

The advantage of the modified algorithm over creating fixed action categories is that actions are automatically grouped together based on their similarity and that the median value is kept as an indication of the values grouped. Note that more information could be kept such as the minimum and maximum values for each cluster, and that a different clustering algorithm could be used.

A test with logs from RomanTutor permitted extracting sequential patterns containing valued actions. One such pattern indicates that learners performed a rotation of joint EP with a median value of 15º, followed by selecting camera 6. Another pattern found consists of applying a rotation of joint EP with a median value of 53º followed by the selection of the camera 4. In this case, the dynamic categorization enhanced the quality of the extracted patterns, since otherwise both patterns would be considered as starting with a "small rotation" of the joint EP.

## 6 Extending the Hirate-Yamana Algorithm to Mine the Compact Representation of Closed Sequences

A third limitation of the Hirate-Yamana algorithm and of the SPM algorithms such as Prefixspan is that among the mined sequences, there can be many redundant sequences. For example, the Hirate-Yamana algorithm could find the three frequent sequences <(0,a)> , <(0,a), (0,b)> and <(0,a), (0,b), (0,c)> in a database. In a tutoring context where we want to extract a task model, only the frequent closed or maximal sequences could be kept, as we are interested by the longer sequences, and this would allow reducing the number of sequences to consider by tutoring services.

"Closed sequences" are sequences that are not contained in another sequence having the same support. A closed pattern induces an equivalence class of pattern sharing the same closure, i.e. all the patterns belonging to the equivalence class are verified by exactly the same set of plans. Those patterns are partially ordered, e.g. considering the inclusion relation. The smallest elements in the equivalence class are called minimal generators, and the unique maximal element is called the closed pattern. On the other hand, "maximal sequences" are sequences that are not contained in any other sequence. In this work, we extend our algorithm to mine closed sequences instead of maximal sequences, since closed sequences are a lossless compact representation of the set of frequent sequences. In other words, the set of closed frequent sequences allows reconstituting the set of all frequent sequences and their support [16] (no information is loss).

We have extended our modified Hirate-Yamana algorithm to find only closed sequences. To achieve this, we have integrated the BI-Directional Extension closure (BIDE) checking of the BIDE+ algorithm [16], which was proposed as an extension of the PrefixSpan algorithm, and permits checking if a sequence is closed without having to maintain a set of closed sequences candidates (as many closed pattern mining algorithm do). The BIDE scheme basically checks if a pattern is closed by checking in the original sequences that contains the pattern if there exist one action with the same support that could extend the pattern. We have also implemented the BackScan pruning of the BIDE+ algorithm that allows stopping growing some sequences that are guaranteed to not produce any closed sequences (see [16] for further information). The BackScan pruning has the advantage of often increasing the time performance over regular SPM algorithms such as PrefixSpan [16]. Because of limited space, the reader is invited to refer to [16] for more details on the BIDE algorithm.

## 7 Extending Sequential Pattern Mining with Context Information

A fourth limitation that we encountered when applying the PrefixSpan algorithm is that it does not consider the context of each sequence. In a tutoring system context, it would be useful, for instance, to annotate sequences with success information and the expertise level of a user and to mine patterns containing this information. Our solution to this issue is to add dimensional information to sequences. Pinto et al. [12] originally proposed Multi-dimensional Sequential Pattern Mining (MDSPM), as an extension to SPM. A Multidimensional-Database (MD-Database) is defined as a sequence database having a set of dimensions $D=\{D_1, D_2,\ldots D_n\}$. Each sequence of a MD-Database (an MD-Sequence) possesses a symbolic value for each dimension. This set of value is called an MD-Pattern and is noted $\{d_1, d_2\ldots d_n\}$. For example, consider the MD-Database depicted in the left part of figure 5. The MD-Sequence 1 has the MD-Pattern {"true", "novice"} for the dimensions "success" and "expertise level". The symbol "*", which means any values, can also be used in an MD-Pattern. This symbol subsumes all other dimension values. An MD-Pattern $P_x=\{d_{x1}, d_{x2}\ldots d_{xn}\}$ is said to be contained in another MD-Pattern $P_y=\{d_{y1}, d_{y2}\ldots d_{ym}\}$ if there exists integers $1 \leq i_1 < i_2 < \ldots i_n \leq m$ such that $d_{x1} \subseteq d_{y1}, d_{x2} \subseteq d_{y2}, \ldots, d_{xn} \subseteq d_{yn}$. The problem of MDSPM is to find all MD-Sequence appearing in a database with a support higher than *minsup*. Figure 5 shows an MD-Database with time information and some patterns that can be extracted from it, with a *minsup* of 2 sequences.

| An MD-Database | | |
|---|---|---|
| **ID** | **Dimensions** | **Sequences** |
| 1 | true, novice | <(0,a),(1,bc)> |
| 2 | true, expert | <(0,d) > |
| 3 | false, novice | <(0,a),(1,bc)> |
| 4 | false, interm. | <(0,a),(1,c), (2,d)> |
| 5 | true, novice | <(0,d), (1,c)> |
| 6 | true, expert | <(0,c), (1,d) |

→

| Mined MD-Sequences | |
|---|---|
| **Dimensions** | **Sequences** |
| *, novice, | <(0,a)> |
| *, * | <(0,a)> |
| *, novice | <(0,a), (1,b)> |
| true, * | <(0,d)> |
| true, novice | <(0,c)> |
| true, expert | <(0,d)> |

**Fig. 5.** An Example of SPM with Dimensions and Time Information

Pinto et al. [17] proposed three algorithms for MDSPM. The first one cannot be applied in combination with the Hirate-Yamana algorithm, as it required embedding dimensions as additional actions in sequences. The two other algorithms, SeqDim and DimSeq are based on the idea that the problem of MDSPM can be broken in two steps: finding sequential patterns with an algorithm such as PrefixSpan, and finding MD-Patterns with an itemset mining algorithm such as Apriori [18, 19]. The first algorithm, SeqDim is executed as follow. First, frequent sequences are found by SPM. Then, for each sequence, the containing MD-Sequences are used to mine frequent MD-Patterns which are then combined with the sequence to form MD-Sequence(s). Alternatively, the second algorithm, DimSeq, first perform a search for frequent MD-Patterns. Then, for each pattern, the containing MD-Sequences are used to mine frequent sequences, which are then combined with the pattern to form MD-Sequence(s).

In our implementation, we chose SeqDim and integrated it with our extended Hirate-Yamana algorithm. For MD-Patterns mining, we applied the AprioriClose algorithm [19], but any itemset mining algorithm such as Apriori [18] can be used. We chose AprioriClose as it allows mining the set of closed MD-Patterns, and thus to eliminate some redundancy among the mined sequences. It is important to note that combining closed MD-Patterns mining and closed SPM does not results in closed MDSPM [20]. In future work we plan to adapt our algorithm as suggested by [20] to achieve closed MDSPM, and remove more redundancy.

Applying our modified algorithm with MDSPM in RomanTutor showed to be useful as it allowed to successfully identify patterns common to all expertise levels that lead to failure ("*, failure"), for example. Currently, we have encoded two dimensions: expertise level and success. But additional dimensions can be easily added. In future work, we plan to encode skills involved as dimensional information (each skill could be encoded as a dimension). This will allow computing a subset of skills that characterize a pattern by finding common skills demonstrated by users who used that pattern. This will allow diagnosing missing and misunderstanding skill for users who demonstrated a pattern.

## 8 Dividing Long Problems into Sub-problems

The last improvement that we made is to how we apply the algorithm to extract a partial problem space. Originally, we mined frequent patterns from sequences of user actions for a whole problem-solving exercise. But, we noticed that in general, after more than 6 actions performed by a learner, it becomes hard for the system to tell which pattern the learner is doing. For this reason, we added the definition of "problem states". For example, in the RomanTutor, where an exercise consists of moving a robotic arm to attain a specific arm configuration, the 3D space was divided into 3D cubes, and the problem state at a given moment is defined as the set of cubes containing the arm joints. An exercise is then viewed as going from a problem state $P_1$ to a problem state $P_F$. For each attempt at solving the exercise, we log (1) the sequence of problem states visited by the learner A={$P_1$, $P_2$... $P_n$} and (2) the list of actions performed by the learner to go from each problem state to the next visited problem state ($P_1$ to $P_2$, $P_2$ to $P_3$, ... $P_{n-1}$ to $P_n$). After many users performed the same exercise, we extract sequential patterns from (1) the sequences of problems states visited, and (2) from the sequences of actions performed for going from a problem state to another.

Dividing long problems into sub-problems allow a better guidance of the learner, because at any moment, only the patterns starting from the current problem state have to be considered.

We describe next how the main tutoring services are implemented. To recognize a learner's plan, the system proceeds as follow. The first action of the learner is compared with the first action of each pattern for the current problem state. The system discards the patterns that do not match. Each time the learner makes an action, the system compares the actions done so far by the learner with the remaining patterns. When the problem-state changes, the system considers the set of patterns associated to the new problem state. If at any given moment a user action does not match with any patterns, the algorithm ignores the last user action or the current action to match for each pattern. This makes the plan recognizing algorithm more flexible and has shown to improve its effectiveness. At a more coarse grain level, a tracking of the problem states visited by the learners is achieved similarly as the tracking for actions.

One utility of the plan recognizing algorithm for actions/problem states is to assess the expertise level of the learner (novice, intermediate or expert) by looking at the patterns applied. The plan recognizing algorithm also allows suggesting to the learner the possible actions from the current state. In RomanTutor, this functionality is triggered when the student selects "What should I do next?" in the interface menu. In this case, the tutoring service selects the action among the set of patterns that has the highest relative support and that is the most appropriate for the estimated expertise level of the learner. When no actions can be identified, RomanTutor relies on a path-planner [6] to generate an approximate path to the goal.

## 9 A Preliminary Experiment

We conducted a preliminary experiment in RomanTutor. We have set up two scenarios consisting each of moving a load with the Canadarm2 robotic arm to one of the two cubes (figure 6). A total of 12 users (a mix of novices, intermediates and experts) have been invited to execute these scenarios using the CanadarmII robot simulator. The number of primitive actions that have been retained is 112 (some have been redefined from the original 155 to have values, thus reducing the number of different possible actions). The expertise levels and success/failure information was added manually to sequences. From this data set, we extracted 558 sequential patterns for problem-states and actions with the extended SPM algorithm. These patterns were then used as input by the tutoring services of RomanTutor. In a subsequent work session, we asked the users to compare the tutoring services offered in our first experiment with RomanTutor with those offered with the newly extracted knowledge base. On the whole, users preferred the newer version, as the hints offered were generally more precise and more appropriate, and help could be provided in more situations. It was also observed that the system more often correctly inferred the estimated expertise level of learners by using the dimensional information.

Figure 7 illustrates a hint message given to a learner upon request during scenario 1. The guiding tutoring service selected the pattern that has the highest support value, matches the last student actions and problem-state, is marked "success" and corresponds with the estimated expertise level of the learner. The given hint is to decrease the rotation value of the joint "EP", increase the rotation value of joint "WY", and finally to select camera "CP2" on "Monitor1". The values on the right column indicate the values associated to the action. In this context, the values "2" and "3" means to rotate the joints 20 º and 30 º, respectively (1 unit equals 10º). By default, three steps are showed to the learners in the hint window depicted in figure 7. However, the learner can click on the "More" button to ask for more steps or click on the "another possibility" button to ask for an alternative.

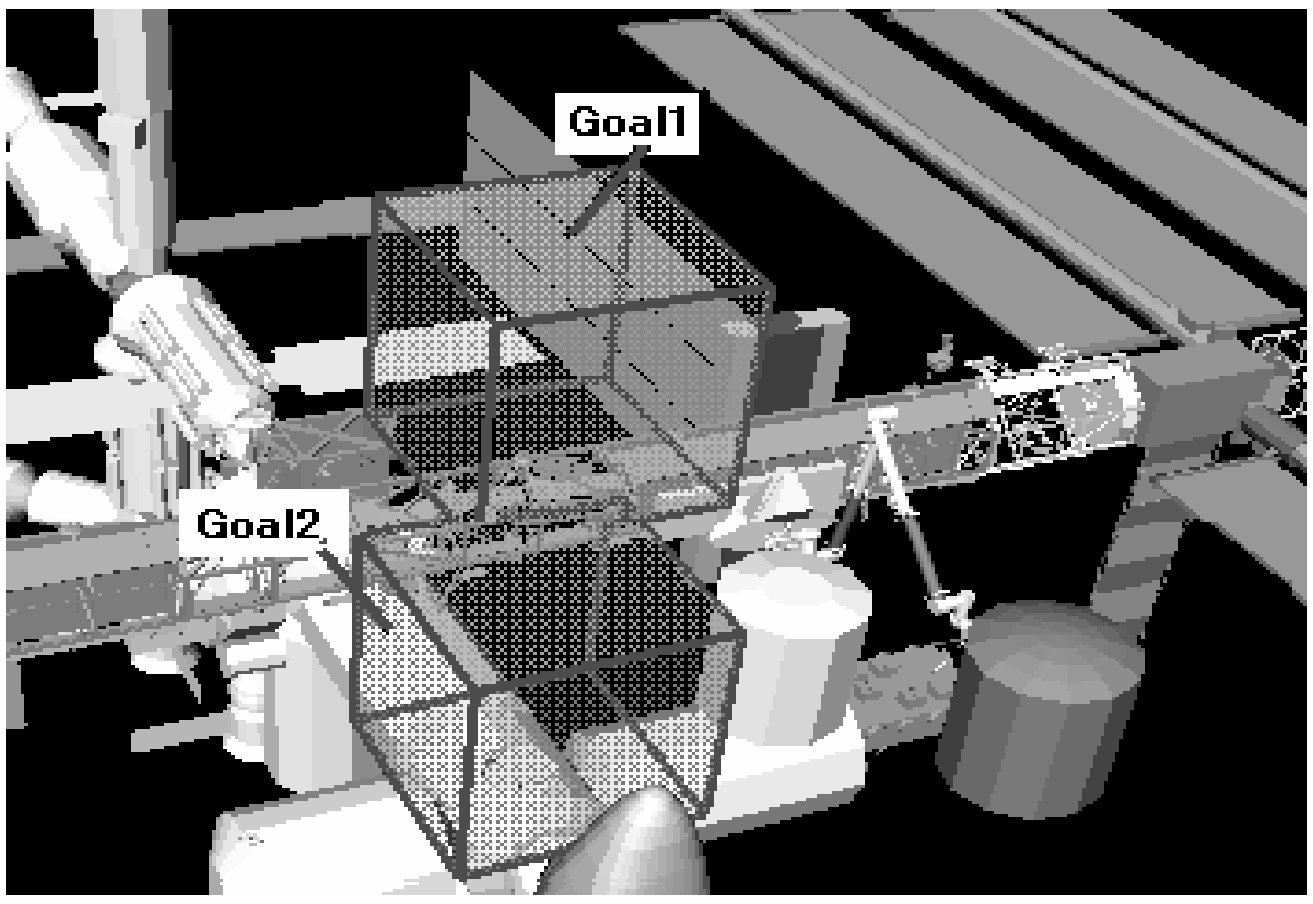

**Fig. 6.** The two manipulation scenarios

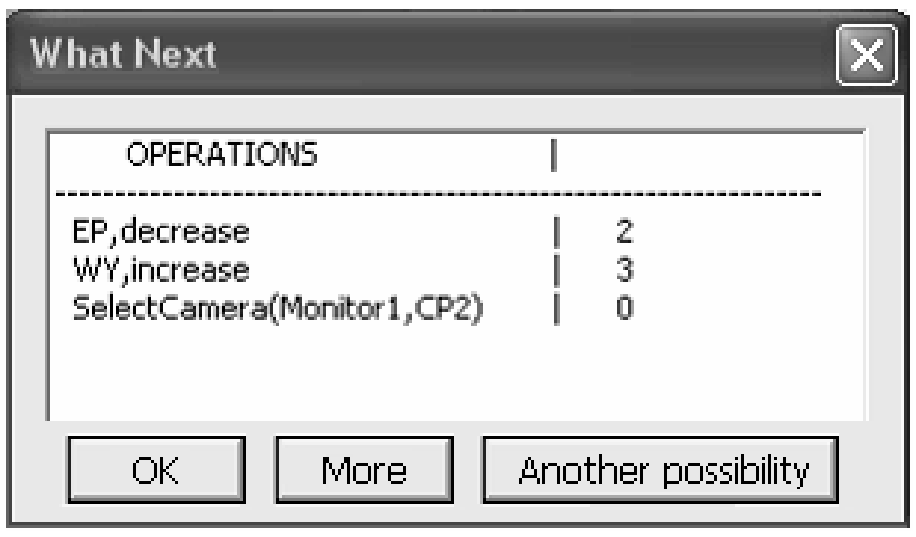

**Fig. 7.** A hint generated by the guiding tutoring service

## 10 Conclusion

Following our first experiment of applying a SPM based framework [5] in RomanTutor, we reported in this paper several limitations of SPM for learning procedural knowledge associated to a task, and proposed a new SPM framework to overcome these limitations. The SPM algorithm combines (1) time intervals, (2) closed sequential pattern mining, (3) multi-dimensional pattern mining and (4) the automatic clustering of valued actions. We also suggested dividing problems into problem states to enhance the relevance of the tutoring services.

The framework was used to extract a problem space, and support tutoring services in RomanTutor. Since the framework proposed in this paper and its inputs and outputs are domain independent, the framework can be potentially applied to any ill-defined procedural domains where the problem can be stated in the same way.

For future work, we are first working on including skills as dimensional information, and on conducting an experiment with a larger group of learners. We are also planning to develop new tutoring services to exploit the problem space.

Finally, we plan to use association rules mining as in our previous work, to find associations between patterns over a whole problem-solving exercise [5]. This could improve the effectiveness of the tutoring services, as it would be complementary to dividing the problem into problem states. For example, if a learner performed a pattern *p*, an association rule could indicate that the learner has a higher probability of applying another pattern *q* later during the exercise than another pattern *r* that is available for the same problem state.

**Acknowledgment.** Our thanks go to the FQRNT and NSERC for their logistic and financial support. The authors would like to thank Severin Vigot and Mikael Watrelot for integrating the framework in RomanTutor, and Khaled Belghith, Daniel Dubois, Usef Faghihi, Mohamed Gaha and the other members of the GDAC/PLANIART teams who participated in the development of RomanTutor.